\documentclass{article}

\usepackage{microtype}
\usepackage{graphicx}
\usepackage{subfigure}
\usepackage{booktabs} 

\usepackage{hyperref}

\usepackage[accepted]{icml2024}

\usepackage{amsmath}
\usepackage{amssymb}
\usepackage{mathtools}
\usepackage{amsthm}
\usepackage{natbib} 
\usepackage{pifont}
\usepackage{multirow}

\usepackage{amsfonts}

\newcommand{\Ab}{\mathbf{A}}
\newcommand{\Bb}{\mathbf{B}}
\newcommand{\Cb}{\mathbf{C}}
\newcommand{\Db}{\mathbf{D}}

\newcommand{\Fb}{\mathbf{F}}
\newcommand{\Gb}{\mathbf{G}}

\newcommand{\Ib}{\mathbf{I}}

\newcommand{\Kb}{\mathbf{K}}

\newcommand{\Mb}{\mathbf{M}}

\newcommand{\Ob}{\mathbf{O}}
\newcommand{\Pb}{\mathbf{P}}
\newcommand{\Qb}{\mathbf{Q}}

\newcommand{\Ub}{\mathbf{U}}
\newcommand{\Vb}{\mathbf{V}}
\newcommand{\Wb}{\mathbf{W}}
\newcommand{\Xb}{\mathbf{X}}
\newcommand{\Yb}{\mathbf{Y}}
\newcommand{\Zb}{\mathbf{Z}}

\newcommand{\RR}{\mathbb{R}}

\newcommand{\xmark}{\ding{55}}%
\newcommand{\ours}{\textsc{CHELA}}

\usepackage[capitalize,noabbrev]{cleveref}

\theoremstyle{plain}

\theoremstyle{definition}

\theoremstyle{remark}

\usepackage[textsize=tiny]{todonotes}

\icmltitlerunning{Short-Long Convolutions Help Hardware-Efficient Linear Attention to Focus on Long Sequences}

\begin{document}

\twocolumn[
\icmltitle{Short-Long Convolutions Help Hardware-Efficient Linear Attention \\ to Focus on Long Sequences}




\begin{icmlauthorlist}
\icmlauthor{Zicheng Liu}{westlake}
\icmlauthor{Siyuan Li}{westlake}
\icmlauthor{Li Wang}{westlake}
\icmlauthor{Zedong Wang}{westlake}
\icmlauthor{Yunfan Liu}{westlake}
\icmlauthor{Stan Z. Li}{westlake}
\end{icmlauthorlist}

\icmlaffiliation{westlake}{AI Lab, Research Center for Industries of the Future, Westlake University, Hangzhou, China}

\icmlcorrespondingauthor{Stan Z. Li}{stan.zq.li@westlake.edu.cn}

\icmlkeywords{Machine Learning, ICML}

\vskip 0.3in
]



\printAffiliationsAndNotice{}  

\begin{abstract}
To mitigate the computational complexity in the self-attention mechanism on long sequences, linear attention utilizes computation tricks to achieve linear complexity, while state space models (SSMs) popularize a favourable practice of using non-data-dependent memory pattern, \textit{i.e.,} emphasize the near and neglect the distant, to processing sequences. Recent studies have shown the priorities by combining them as one. However, the efficiency of linear attention remains only at the theoretical level in a causal setting, and SSMs require various designed constraints to operate effectively on specific data. Therefore, in order to unveil the true power of the hybrid design, the following two issues need to be addressed: (1) hardware-efficient implementation for linear attention and (2) stabilization of SSMs. To achieve this, we leverage the thought of tiling and hierarchy to propose \textsc{CHELA} (short-long \textbf{C}onvolutions with \textbf{H}ardware-\textbf{E}fficient \textbf{L}inear \textbf{A}ttention), which replaces SSMs with short-long convolutions and implements linear attention in a divide-and-conquer manner. This approach enjoys global abstraction and data-dependent selection from stable SSM and linear attention while maintaining real linear complexity. Our comprehensive experiments on the Long Range Arena benchmark and language modeling tasks demonstrate the effectiveness of the proposed method.

\end{abstract}

\section{Introduction}
\label{sec:intro}
\begin{figure*}
    \centering
    \includegraphics[width=0.85\linewidth]{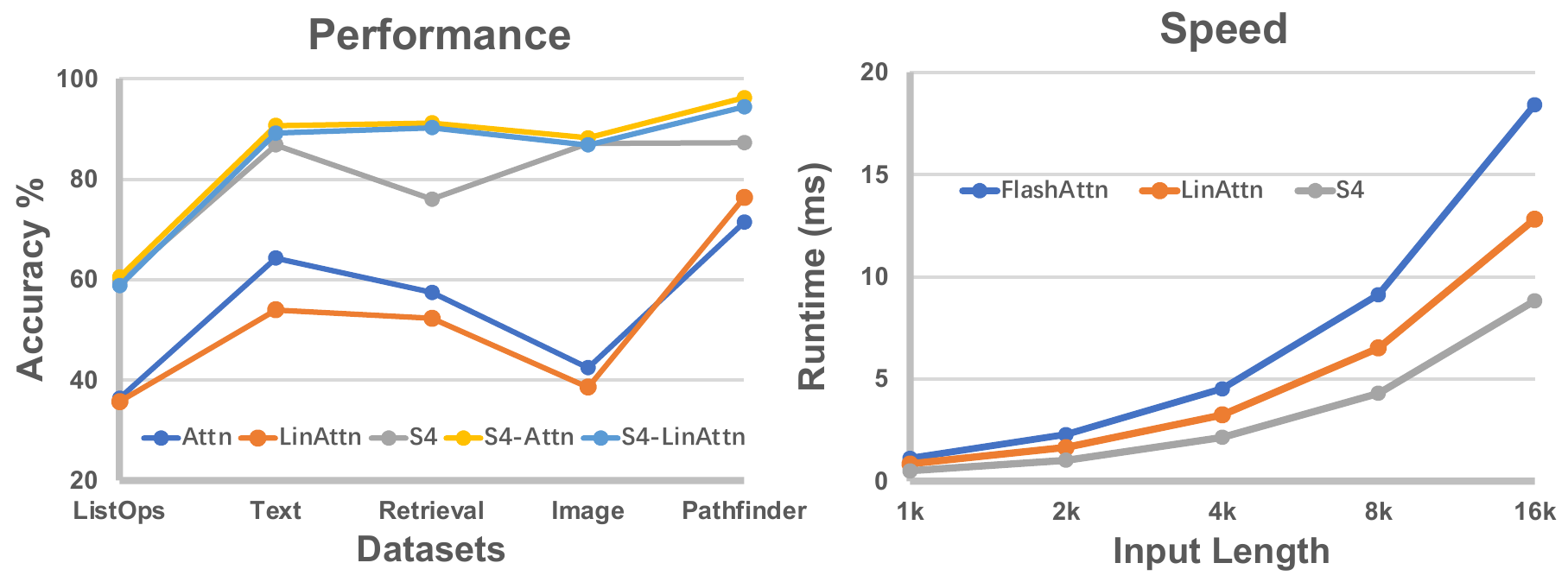}
    \caption{Demonstration of (\textit{left}) comparison of various popular models on the Long Range Arena Dataset, and (\textit{right}) speed benchmark on different implementations of attentions. Figure \textit{left}, despite showing promising performance of SSM-Attention models, the linear version of these models degenerated. Figure \textit{right} benchmarks attention speed with around 200M parameters, showing linear attention is not linear with sequence length and is significantly slower than S4, which can be a speed bottleneck in these hybrid designs.}
    \label{fig:intro}
\end{figure*}

Transformer models have demonstrated remarkable performance on a range of natural language processing tasks~\citep{vaswani2017attention}, such as language modeling~\citep{devlin2018bert}, visual signal processing~\citep{dosovitskiy2021vit,liu2022automix,li2023moganet,liu2023decoupledmix}, and speech understanding~\citep{gulati2020conformer}. These models use the attention mechanism, which calculates a dependency score for each pair of tokens in an input sequence. Consequently, full attention has a quadratic time and space complexity relative to the sequence length. This complexity, however, becomes computationally prohibitive for tasks that involve long sequences~\citep{lin2022survey}.
It is worth mentioning that Transformer models equipped with full attention tend to overfit. This is because the attention mechanism does not make any assumptions about the structure of the inputs, which leads to the absence of structural biases. To train a Transformer model, even the order information has to be included. Therefore, the full attention is too flexible to overfit to noise. This limitation restricts the practicality of these models in long sequence modeling, where the dependency signal is often weak and the signal-to-noise ratio is low.
To solve this, recent studies have designed hybrid models~\citep{ma2022mega,zuo2023spade} by combining efficient state space models (SSMs)~\citep{gu2021efficiently,gu2020hippo,gu2022parameterization,hasani2022liquid,smith2023simplified}, with expressive attention variants for modeling long sequences from perspectives in structured and flexible patterns, achieving promising results. It is worth noting that their complexity is essentially quadratic, and the corresponding linear versions both suffer performance degradation (see Fig.~\ref{fig:intro} \textit{left}).

While it is true that the simple use of a chunk linearization strategy can rival the speed of SSMs, the drop in performance is significant. We, therefore, abandon this strategy and the softmax function to accelerate the linear attention operations of the kernel-based approach. In addition, long convolution~\citep{fu2023longconv} is more efficient and easier to implement than the traditional SSMs. We believe combining these two modules could release the true power of the hybrid model with faster speed and better performance. However, the existence of such a dilemma is worth considering: \textbf{(1) the linear attention in hybrid models is a speed bottleneck for long convolution, the comparison is shown in Fig.~\ref{fig:intro} \textit{right}; (2) the instability of long convolution makes attention hard to optimize.}
Specifically, due to the intense memory access (I/O) on the GPU and cumulative summation (\texttt{cumsum}) operation in a casual setting, the notable reduction in complexity from $\mathcal{O}(L^2)$ to $\mathcal{O}(L)$ in linear attention is only theoretical~\citep{qin2024transnormerllm}; On the other hand, the long convolution needs hand-crafted regularizations to prevent over-fitting high-frequency noise. 

To escape the dilemma, we have developed a novel model called {\ours}, which stands for short-long \underline{\textbf{C}}onvolutional with \underline{\textbf{H}}ardware-\underline{\textbf{E}}fficient \underline{\textbf{L}}inear \underline{\textbf{A}}ttention. This model is designed to effectively and efficiently capture complex dependencies in long inputs. Inspired by FlashAttention~\citep{dao2022flashattention1,dao2023flashattention2}, the model comprises multiple layers of real linear token mixers with hardware-friendly implementation with a gating mechanism that achieves comparable performance to full attention. We have added a stabilizer, reparameterizable~\citep{ding2022replknet,ding2023unireplknet} short-long convolutions, to the bottom layer of the long convolution to integrate multiple frequencies with inputs. As a result, the top layers of {\ours} capture more refined data-dependent information using linear attention, while the convolutions add a structural bias that enhances global abstraction in multi-level frequencies.
We demonstrate the efficiency and effectiveness of {\ours} on various datasets and tasks.
First, we show that the proposed method outperforms existing approaches on the Long Range Arena (LRA) benchmark~\cite{tay2020long}, which is designed to test models' ability in modeling long sequences.
Second, we show that in autoregressive language modeling, {\ours} is not only significantly faster than the vanilla Transformer but also yields better performance.
In all the settings, {\ours} outperforms the baselines.
Finally, we provide further analysis and ablation experiments to demonstrate the effectiveness.

Our contribution can be summarized in three folds:
\begin{itemize}
    \item \textbf{Efficient Implementation}: We release the potential speed of linear attention in the attention-ssm hybrid model, maintaining a constant rate regardless of sequence length with fixed memory.
    \item \textbf{New Design}: We improve multi-frequency learning with multi-level convolutional hierarchies, leading to better performance and stability for long convolution.
    \item \textbf{Promising Results:} By connecting the above two basic modules, we propose {\ours} for efficient long sequence learners are capable of both global information abstraction and fine-grained data-dependent selection.
\end{itemize}

\section{Background}
\subsection{Self Attention}
\label{sec.attention}

If we have an input $\Xb$ that belongs to $\RR^{L \times d}$, where $L$ represents sequence length and $d$ represents the embedding dimension, the attention mechanism produces pair-wise scores denoted as $\Mb$:
\begin{align} \label{eq:attention}
    &\Mb = \mathrm{Attn}( \Qb, \Kb, \Vb ) = \sigma \left ( \frac{ \Qb \Kb^\top }{\sqrt{d}} \right ) \Vb, \\
    &\text{where } \Qb = \Xb \Wb_q, \ \Kb = \Xb \Wb_k, \ \Vb = \Xb \Wb_v. \notag
\end{align}
Here $\Wb_q, \Wb_k, \Wb_v \in \RR^{d\times d}$ are learnable weights, and $\sigma$ is the attention function. Denote the attention matrix $\Mb \in \RR^{L \times L}$. $\Mb_{ij}$ captures the relations between the $i$-th and the $j$-th input tokens.
\paragraph{Linear attention}
Taking Norm as $\sigma$~\citep{qin2024transnormerllm}, to take advantage of the computational efficiency inherent in right matrix multiplication, the attention score $\Mb$ can be transformed into its linear variant, which is mathematically equivalent due to the properties of matrix multiplication.
\begin{equation}
    \Mb_{\texttt{linear}} = \texttt{Norm}(\Qb(\Kb^\top \Vb))
\end{equation}


\subsection{Linear Recurrent Models}

\textbf{Continuous formulation.}
A model with a continuous time latent space transforms a one-dimensional input signal $u(t)$ into a latent state $x(t)$ that has $d_s$ dimensions, and then this $x(t)$ is transformed into a one-dimensional output signal $y(t)$. Clearly, we can define the process:
\begin{align} \label{eq:ssm}
    x'(t) = \Ab x( t)  + \Bb u( t ), \quad y( t ) = \Cb x( t ).
\end{align}
where $\Ab \in \RR^{ d_s \times d_s }$, $\Bb \in \RR^{ d_s }$ and $\Cb \in \RR^{ d_s }$.

Eq.~\ref{eq:ssm} is utilized in prior research to model long sequences. For instance, \citet{gu2020hippo} suggests a set of matrices called HiPPO (high-order polynomial projection operators) to initialize $\Ab$. The HiPPO matrices are constructed to enable the state $x(t)$ at time $t$ to remember the history of the input $u(t)$ up to the present.


\paragraph{Discrete formulation.}
Discrete sequences, like natural language inputs $(u_0, u_1, \cdots, u_L)$ with $L$ representing sequence length, are frequently encountered. The model described in Eq.~\ref{eq:ssm} can be discretized (bilinear method, for example) with a step size of $\Delta$ to model discrete data.
\begin{align}
    &x_k = \overline{ \Ab } x_{ k-1 } + \overline{ \Bb } u_{ k }, \quad y_k = \overline{ \Cb } x_k, \\
    &\text{where }\overline{ \Ab } = ( \Ib - \Delta/2 \cdot \Ab )^{ -1 } ( \Ib + \Delta/2 \cdot \Ab ), \notag \\ 
    &\quad\quad\ \ \ \overline{ \Bb } = ( \Ib - \Delta/2 \cdot \Ab )^{-1} \Delta \Bb, \quad \overline{ \Cb } = \Cb. \notag
\end{align}
We can expand the recurrent representation above to obtain:
\begin{align*}
    y_k = \overline{ \Cb } \overline{ \Ab }^k \overline{ \Bb } u_0 + \cdots + \overline{ \Cb } \overline{ \Ab } \overline{ \Bb } u_{k-1} + \overline{ \Cb } \overline{ \Bb } u_k.
\end{align*}
\paragraph{Convolution formulation.} This can be written as a convolutional representation $y = \overline{ \Kb } * u$, where the convolution kernel
\begin{align} \label{eq:ssm-conv}
    & \overline{ \Kb } \in \RR^L = \left( \overline{ \Cb } \overline{ \Bb }, \overline{ \Cb } \overline{ \Ab } \overline{ \Bb }, \cdots, \overline{ \Cb } \overline{ \Ab }^{L-1} \overline{ \Bb } \right ).
\end{align}
Efficient calculation of the output $y$ in Eq.~\ref{eq:ssm-conv} can be achieved when the convolution kernel $\overline{K}$ is known. However, determining the kernel is a difficult undertaking, and the majority of existing algorithms necessitate a time and space complexity of $O(L^2)$.

\paragraph{S4 model.}
\label{sec.longconv}
\citet{gu2021efficiently} devised the S4 model to effectively calculate Eq.~\ref{eq:ssm-conv}. In particular, the initialization of $\Cb$ in Eq.~\ref{eq:ssm} is random, and both $\Ab$ and $\Bb$ are initialized as
\begin{align}
    &\Ab = \Ab^{ (d_s) } - \Pb \Pb^\top, \quad \Bb_i = (2i + 1)^{ \frac{1}{2} }, \\
    &\text{where } \Pb_i = \left( i + 1/2 \right)^{1/2}, \notag \\
    &\Ab^{ (d_s) }_{ ij } = -
    \begin{cases}
        (i + \frac{ 1 }{ 2 })^{\frac{ 1 }{ 2 }} (j + \frac{ 1 }{ 2 })^{\frac{ 1 }{ 2 }}, & i > j, \\
        \frac{ 1 }{ 2 }, & i = j, \\
        -(i + \frac{ 1 }{ 2 })^{\frac{ 1 }{ 2 } } (j + \frac{ 1 }{ 2 })^{\frac{ 1 }{ 2 } }, & i < j. \\
    \end{cases} \notag
\end{align}
The convolution kernel $\overline{\Kb}$ in Eq.~\ref{eq:ssm-conv} can be efficiently computed using $O(L)$ time and space complexity.
After that, the efficient computation of $y=\overline{\Kb} * u$ for an input $u$ can be performed.
\paragraph{Long convolution.} If we parameterize the kernel with the same length as the sequence in Eq.~\ref{eq:ssm-conv}~\citep{fu2023longconv}, then we can replace the SSM layer with a learned convolution kernel as a drop-in replacement. The computation can be efficiently done by using the FFT theorem:
\begin{equation}
    y=u*\Kb=\Fb_L^{-1}\Db_{\Kb}\Fb_{L}u
\end{equation}
where $\Fb_L$ denotes the DFT matrix of size $L$, and $D_K=\mathrm{diag}(\Fb_LK)$. This so-called FFT convolution scales in linear complexity when dealing with sequences with length $L$.


\begin{figure*}[t!]
    \centering
    \includegraphics[width=0.88\linewidth]{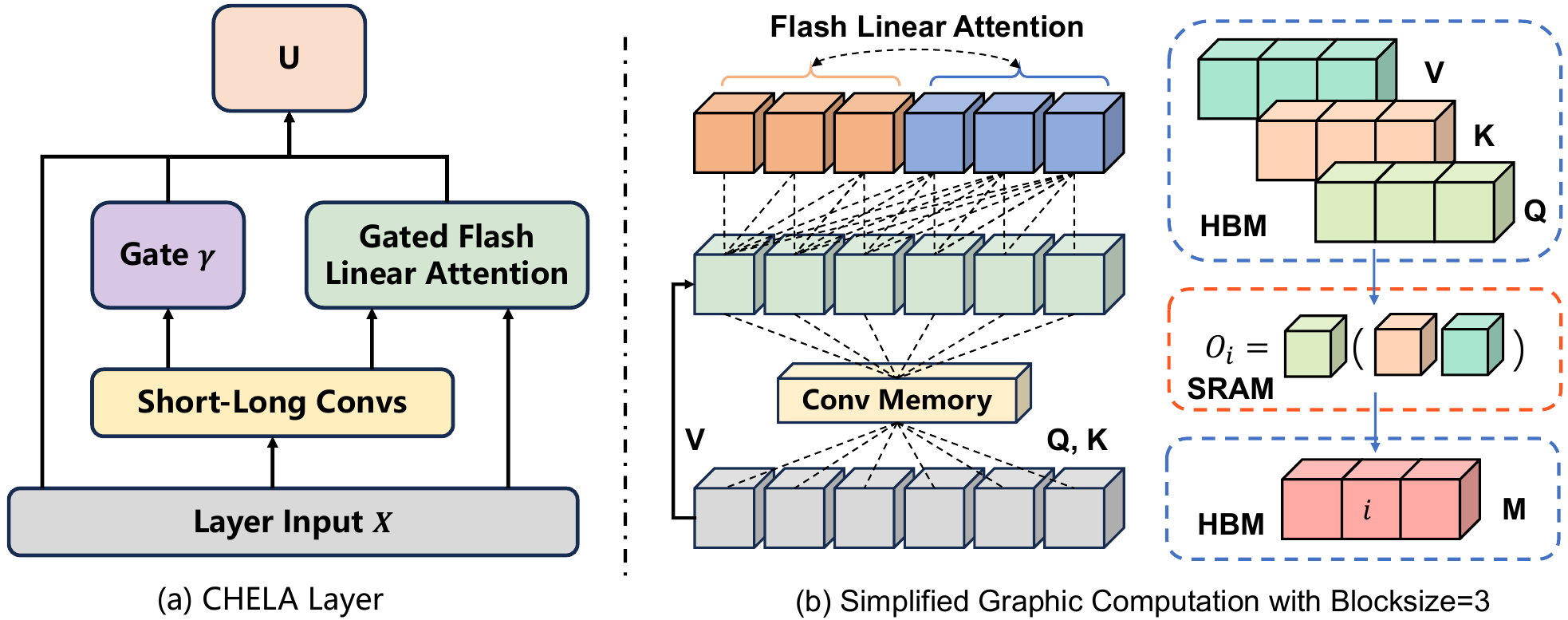}
    \caption{{\ours}-model architecture. Figure (a) shows the {\ours} layer. Figure (b) is a simplified graphic demonstration of the key components of {\ours}: Short-Long Convolutions and Flash Linear Attention. HBM and SRAM denote high bandwidth memory and static random-access memory in GPU, respectively.}
    \label{fig:arch}
\end{figure*}

\section{Why Combining Long Conv and Attention?}
As mentioned in Section \ref{sec:intro}, both long convolution and linear attention mechanisms have their limitations in spite of their widespread use and impressive accomplishments in sequence modeling. By bridging their benefits together, we push the limitation of the hybrid model, which allows us to benefit from the strong inductive bias of structured memory while still having the ability to learn complex data-dependent patterns in true linear time with respect to length. The following two issues are the two main obstacles to extreme performance in long-sequence modeling.

\paragraph{Linear Attention is the Efficiency Bottleneck.}
We can see from Fig.~\ref{fig:intro} \textit{right} that as the input sequence grows, the actual efficiency of linear attention is much lower than that of S4, despite the fact that the complexity of the S4 model is $\mathcal{O}(L\log L)$ as for long convolution and $\mathcal{O}(L)$ for linear attention. The two obstacles to achieving a practical wall-clock speedup are intensive GPU memory access and cumulative summation (\texttt{cumsum}) in a casual setting. Inspired by Flashattention~\cite{dao2022flashattention1,dao2023flashattention2}, we effectively doubled the speed of linear attention.

\paragraph{Long Convolution Needs Short Convolutions.}
To match the efficiency of hardware-efficient linear attention, the long convolution is considered as a simpler and faster SSM model for long sequence abstraction. However, it requires additional constraint regularizations for stable performance on different data types. This is because it is hard for a single long convolutional kernel to learn multiple frequencies simultaneously, \textit{i.e.,} token patterns that repeat a lot versus those that occur more sparsely. We thereby introduce multi-resolution short convolutional kernels to share the learning pressure of long convolution. They can also be reparameterized by simple linear transformation into a single kernel to speed up the inference phase.

\section{\textsc{CHELA}: Short-Long Convolutions with Hardware-Efficient Linear Attention}
In this section, motivated by the above issues, we propose CHELA, short-long convolutions with hardware-efficient linear attention, which fully enjoys the benefits of each. We first describe how the tiling method is used in linear attention to reach the theoretical performance. We then introduce a simple and effective module named short-long convolutions to enhance the ability of global abstraction in this hybrid design. The blocks of CHELA are demonstrated in detailed architecture, including feed-forward and normalization layers. Moreover, we also discuss the relationship between {\ours} and two closely related hybrid models: MEGA~\citep{ma2022mega}, SPADE~\citep{zuo2023spade}.

\begin{figure}[b!]
    \centering
    \includegraphics[width=0.95\linewidth]{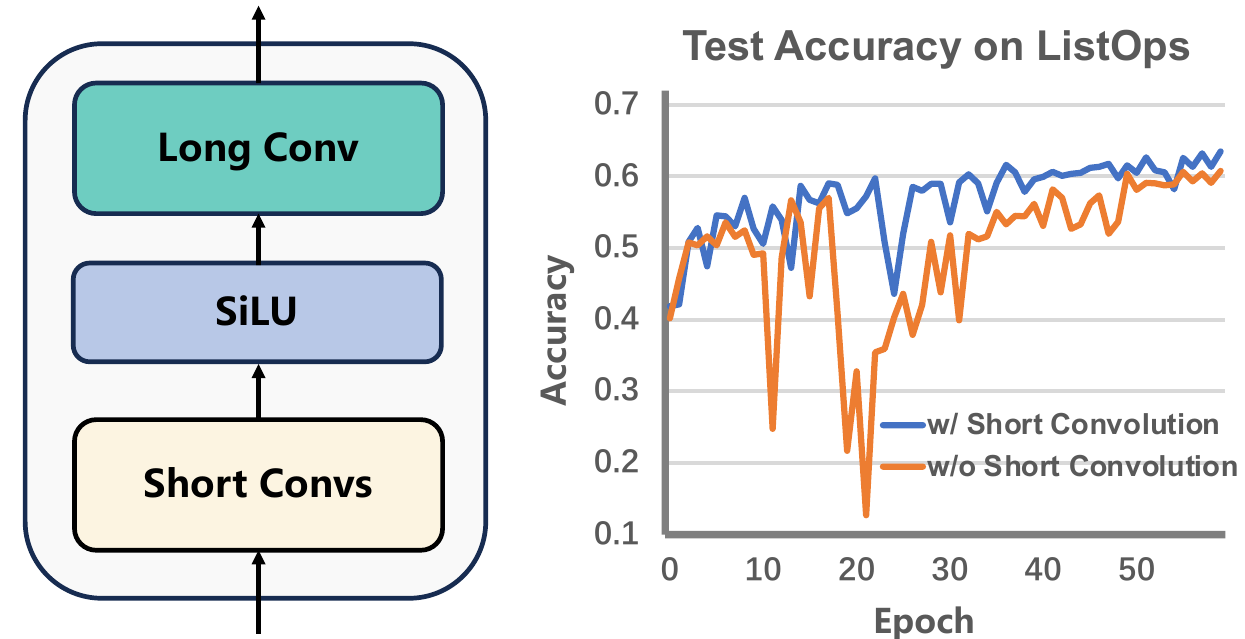}
    \caption{Illustration Short-long Module and the effect of Short Convolution. Figure \textit{left} shows the details of the module structure. Figure \textit{right} proves the role of the short kernel.}
    \label{fig:short_long}
\end{figure}

\subsection{Hardware-Efficient Linear Attention}
In response to the first question, we follow a tiling approach from GLA~\cite{yang2024gla}, which employs the classic divide-and-conquer strategy to take advantage of the significant difference in memory bandwidth between HBM and SRAM within GPU. Given the inputs $\Qb, \Kb, \Vb$, we aim to compute the linear attention in SRAM and finally output $\Ob$. The right-hand multiplication of linear attention is used for the loops between and in the data blocks; noticeably, the left multiplication is used inside the block when an attention mask exists. The intermediate variable $\Ib$ is iteratively saved and accumulated within SRAM. Subsequently, the outputs of outer and inner loops are summed within SRAM, and the results are returned to HBM. This method aims to capitalize on the distinct advantages of each memory component, optimizing the computational workflow of linear attention to reach its theoretical speed as much as possible. The graphic demonstration is visualized in Fig.~\ref{fig:arch} (b).

\subsection{Short-Long Convolutions}
{\ours} introduces a modification of the long convolution mentioned in Sec.~\ref{sec.longconv} by adding short convolutions before the long kernel, named \textit{short-long convolutions}, to learn multi-frequencies information and improve its stability and capacity. Structural Reparameterization (SR) can fuse these short kernels into a single kernel in the inference phase.  

\paragraph{Short convolution improves long convolution.}
The long convolutions have shown success when dealing with long sequences and are much more efficient than conventional SSMs. Empirical results tell us how the weights are initialized and how the regular terms are designed, all of which greatly impact the final performance on specific data types. One important finding is that long convolutions trying to learn both low and high frequencies in the input signal are the central cause of their instability because their kernel mixes various high and low-frequency patterns compared to S4~\citep{fu2023longconv}.
Inspired by this, we try adding parallel short kernels in long convolution. As shown in Fig~\ref{fig:short_long}, we simply put the short kernels and a SiLU activation function $\phi_{\mathrm{silu}}$~\citep{ramachandran2017searching} before the long convolution. Formally, the computation can be written as:
\begin{equation}
    \Zb = \overline{\Kb}_{l}(\phi_{\mathrm{silu}}(\overline{\Kb}_{s}(\Xb)))
\end{equation}
where $\overline{\Kb}_{l}$ and $\overline{\Kb}_{s}$ denote for long and short convolutions. The Fig.~\ref{fig:short_long} \textit{right} also demonstrates the performance of this design on the ListOps dataset: better stabilization and accuracy on the task of long-sequence logic inference. In our experiments, we add two different sizes of short kernels. One is fixed the size at three, and the other varies with the sequence length, whose size equals $2\log_{10}L+1$.

\paragraph{Structural reparamterization.}
Structural Reparameterization~\cite{ding2022replknet,ding2023unireplknet} is a methodology of equivalently converting model structures via linear parameter transformation. For example, RepVGG targeted a deep inference-time VGG-like (e.g., branch-free) model and constructed extra ResNet-style shortcuts parallel. After training, these residual kernels are absorbed into a single kernel of size 3. In this paper, we use this methodology to fuse short kernels after training. Consequently, we make the long convolution kernel capable of capturing flexible patterns, improving overall performance.

\subsection{CHELA Model}
The gating mechanism in {\ours} is based on GRU~\citep{cho2014properties} and GAU~\citep{hua2022transformer} as the backbone architectures, with the output of short-long convolutions layer $\Zb$ embedded into the linear attention matrix. $\Zb$ can be considered a contextual abstraction because it provides global information through multi-level structured patterns:
\begin{align}
    \Mb_\mathrm{linear} &= \mathrm{Norm}(\Qb (\Kb^\top \Vb)) \odot \Gb_a \\
    \Gb_a &= \phi_{\mathrm{silu}}(\Zb \Wb_v + b_g) \\
    \Qb &= \alpha_q \odot \Zb + \beta_q \\ 
    \Kb &= \alpha_k \odot \Zb + \beta_k \\
    \Vb &= \phi_{\mathrm{silu}}(\Xb \Wb_v + b_v)
\end{align}
where $\alpha_q, \alpha_k, \beta_q, \beta_k \in \mathbb{R}^d$ are the learnable scalers and offsets of queries and keys like GAU, respectively. The $\Mb_{\mathrm{linear}}$ is the output of flash linear attention with gate $\Gb_a$. Moreover, we employ an output gate $\Gb_o$ to update final activation output $\Ub$ in a residual style:
\begin{align}
    \Gb_o & = \phi_{\mathrm{sigmoid}}(\Zb \Wb_o + b_o) \\
    \Ub   & =  \Mb_\mathrm{linear} \odot \Gb_o + \Xb \odot (1-\Gb_o)
\end{align}
the architecture of $\ours$ is demonstrated in Fig.~\ref{fig:arch} \textit{(a)}.

\paragraph{$\ours$ Block}
Besides, each standard $\ours$ block is also equipped with a two-layer feedforward neural network, layer normalization \citep{ba2016layer}, and residual connections \citep{he2016deep} as channel mixing. Concretely, given the input $\Xb$ to the $\ours$ block, we have the output $\Yb$ as:
\begin{align*}
    &\Xb_a = \mathrm{CHELA} \left( \mathrm{LayerNorm}(\Xb) \right) \\
    &\Yb = \mathrm{FFN} \left( \mathrm{LayerNorm}(\Xb_a) \right) + \Xb_a.
\end{align*}
Note that here we apply pre-layer normalization, similar to \citet{vaswani2017attention}. There are other works (e.g., \citealt{devlin2018bert}) that apply post-layer normalization, where the normalization is applied after the residual connection.

\subsection{Relation to MEGA and SPADE.}
As hybrid models of SSM and attentions, we adhered to the idea of proposing an efficient variant to utilize both well-structured and fully data-dependent patterns on long-sequence modeling. However, it is undeniable that the linear versions of Mage-chunk and SPADE-chunk have caused significant performance degradation, which means they struggle to make good tradeoffs between complexity and performance. The EMA Mega uses is too simple to capture higher-order information, so the overall performance is heavily dependent on attention. SPADE uses a parallel structure that directly combines the SSM and Attention modules, which reduces the dependence on attention, but the linear strategy of dividing the chunks still limits the model's performance. CHELA maximizes efficiency by enhancing structured representations with short-length convolutions and lossless use of linear attention in a hardware-optimized implementation. 
\textit{Ultimately, \textsc{CHELA} boosting performance while also dramatically increasing speed in linear time.}
\begin{table*}[t] 
    \centering
    \caption{Performance of predicting outcomes of list operations in the LRA~\citep{tay2020long}.  Bold indicates the best-performing model and underlines the second best. Results are taken from either the citation. The training speed and peak memory consumption comparison on the Text task with an input length of 4K.}
    \begin{small}
        \begin{tabular}{l | c c c c c c | c c}
        \toprule
        \textbf{Models}                                     & \textbf{ListOps}                   & \textbf{Text}  & \textbf{Retrieval} & \textbf{Image} & \textbf{Pathfinder} & \textbf{PathX} & \textbf{Avg.}  & \textbf{Speed}       \\ \hline
        \emph{Attention}:                                   &                                    &       &           &       &            &       &       &                   \\
        Transformer \citep{vaswani2017attention}            & 36.37                              & 64.27 & 57.46     & 42.44 & 71.40      & \xmark     & 54.39 & 1.0$\times$  \\
        Local Attention \citep{tay2020long}                 & 15.82                              & 63.98 & 52.98     & 41.46 & 66.63      & \xmark     & 46.06 & 5.3$\times$  \\
        Linear Trans. \citep{katharopoulos2020transformers} & 16.13                              & 65.90 & 53.09     & 42.34 & 75.30      & \xmark     & 50.55 & 4.7$\times$  \\
        Linformer \citep{wang2020linformer}                 & 35.70                              & 53.94 & 52.27     & 38.56 & 76.34      & \xmark     & 51.36 & 5.5$\times$  \\
        Sparse Transformer \citep{child1904generating}      & 17.07                              & 63.58 & 59.59     & 44.24 & 71.71      & \xmark     & 51.24 & 4.2$\times$  \\
        Performer  \citep{choromanski2020rethinking}        & 18.01                              & 65.40 & 53.82     & 42.77 & 77.05      & \xmark     & 51.41 & 5.7$\times$  \\
        Sinkhorn Transformer \citep{tay2020sparse}          & 33.67                              & 61.20 & 53.83     & 41.23 & 67.45      & \xmark     & 51.39 & 3.8$\times$  \\
        Longformer \citep{beltagy2020longformer}            & 35.63                              & 64.02 & 59.29     & 40.83 & 74.87      & \xmark     & 55.01 & 1.1$\times$  \\
        BigBird \citep{zaheer2020big}                       & 36.05                              & 64.02 & 59.29     & 40.83 & 74.87      & \xmark     & 55.01 & 1.1$\times$  \\
        Luna-256 \citep{ma2021luna}                          & 37.25                              & 65.78 & 79.56     & 47.86 & 78.55      & \xmark     & 61.95 & 4.9$\times$  \\
        Reformer \citep{kitaev2020reformer}                 & 37.27                              & 56.10 & 53.40     & 38.07 & 68.50      & \xmark     & 50.67 & 0.8$\times$  \\ \hline
        \emph{State Space Models}:                          &                                    &       &           &       &            &       &       &                        \\
        S4 \citep{gu2022parameterization}                   & 59.60                              & 86.82 & 76.02     & 87.09 & 87.26      & 86.26 & 80.50 & -                      \\
        DSS \citep{gupta2022diagonal}                       & 57.60                               & 59.60 & 86.82     & \underline{90.90} & 88.65      & 94.20 & 86.09 & 4.8$\times$ \\
        S4D \citep{gu2022parameterization}                  & 60.18                              & 87.34 & 91.09     & 87.83 & 93.78      & 92.80 & 85.50 & -                 \\ 
        Mamba \citep{gu2023mamba}                  & 38.02                              & 82.98 & 72.14     & 87.83 & 69.82      & 67.32 & 66.59 & -                 \\ \hline
        \emph{Linear Hybrid}:                                 &                                    &       &           &       &            &       &       &                     \\
        Mega-chunk \citep{ma2022mega}                & 58.76                              & 90.19 & 90.97     & 85.80 & 94.41      & 93.81 & 85.66 & 5.5$\times$  \\
        SPADE-chunk \citep{zuo2023spade}                         & 60.50                              & 90.69 & 91.17     & 88.22 & \underline{96.23}      & 97.60 & 87.40 & 5.5$\times$ \\
        {\ours} (Ours)                                         & \textbf{60.69}    & \textbf{91.10} & \textbf{91.65}     & \textbf{91.12} & \textbf{96.40}      & \textbf{98.15} & \textbf{88.19} & \textbf{5.8$\times$}  \\ \bottomrule
        \end{tabular}
    \end{small}
    \label{tab:lra}
\end{table*}
\section{Experiments}
To assess $\ours$, we performed tests on five standard sequence modeling tasks involving different types of data, and we compared them with the latest cutting-edge models for each task. All experiments were realized based on NVIDIA A100-80G and Pytorch. We used float32 parameters, with bfloat16 precision for most computations. 

\subsection{Long-Context Sequence Modeling}
We conducted experiments to evaluate sequence models using the Long Range Arena (LRA) benchmark. This benchmark, introduced by \citep{tay2020long}, is designed to assess the performance of sequence models in long-context learning. The benchmark includes six tasks, namely ListOps \citep{nangia2018listops}, byte-level Text classification~\citep{maas2011learning}, byte-level document Retrieval~\citep{radev2013acl}, Image classification on sequences of pixels~\citep{krizhevsky2009learning}, Pathfinder~\citep{linsley2018learning}, and its extreme long version (Path-X; \citet{tay2020long}). The input sequences for these tasks range from 1K to 16K tokens and cover various data modalities.
In Table~\ref{tab:lra}, {\ours} is compared to various baselines, such as Transformer and its efficient versions, as well as the top-performing S4 models.
In order to make a fair comparison, we make sure that Mega and S4 have a similar number of parameters by balancing the number of layers and model sizes for each task. The results are based on the average of five runs with different random seeds, and you can find the tuning information and model details in the Appendix.
The performance of our model has been outstanding across all six tasks, achieving an average accuracy of 88.26\% and surpassing all the other comparison methods. Additionally, we assessed the speed of our model when applied to the byte-level classification task with a 4K input. Our hardware-efficient linear mechanism has demonstrated remarkable efficiency, with a speed that is 5.8 times faster. It is important to highlight that our model, with its unique short-long convolutions hybrid design, exhibits even greater efficiency compared to a variety of linear Transformers, Structured State Space Models, and recent hybrid models.

\begin{table}[b]
    \centering
    \vspace{-2em}
    \caption{Accuracy on Speech Commands dataset.}
    \label{tab:sc}
    \begin{small}
        \begin{tabular}{l|cc}
        \toprule
                    & \multicolumn{2}{c}{\textbf{SpeechCommand-Raw}} \\
        \textbf{Model}       & \textbf{\#Param.}          & \textbf{Accuracy}          \\ \hline
        Transformer & 786K              &  \xmark                 \\
        S4~\citep{gu2021efficiently}     & 300K              & \underline{97.50}             \\
        Mega~\citep{ma2022mega}  & -              & \xmark             \\
        Mega-chunk~\citep{ma2022mega}  & 476K              & 96.03             \\
        {\ours} (ours)     & 493K              & \textbf{97.98}             \\ \bottomrule
        \end{tabular}
    \end{small}
\end{table}

\subsection{Raw Speech Classification}
We intend to evaluate the capability of {\ours} in modeling lengthy speech signals by employing it for the classification of unaltered speech signals with a duration of 16000, instead of depending on traditional preprocessing techniques like converting them into MFCC features. As per \citet{gu2021efficiently} approach, we classify speech on the Speech Commands dataset's SC10 subset, which was introduced by \citet{warden2018speech}. As reported in \citep{ma2022mega}, the Mega-chunk uses a chunk size of 1000 to enable processing the data. 
In Table~\ref{tab:sc}, our model has 493K parameters and achieves a 97.98\% accuracy, making it the leading method in this table. This result is primarily due to the suitability of long convolutions for processing the numerous continuous and low-frequency signals present in speech. Additionally, the ability of short convolutions to capture rich global information enables attention to focus on important aspects.

\begin{table}[t]
    \centering
    \caption{Performance of pixel-level classification on the sCIFAR.}
    \label{tab:scifar}
        \begin{small}
            \begin{tabular}{lccc}
                \toprule
                \textbf{Model} & \textbf{Accuracy} (\%) \\
                \midrule
                \emph{Attention}: & \\
                Transformer \citep{trinh2018learning} & 62.20 \\
                \midrule
                \emph{RNN}: & \\
                LSTM \citep{hochreiter1997long} & 63.01 \\
                r-LSTM \citep{trinh2018learning} & 72.20 \\
                UR-GRU \citep{gu2020improving} & 74.40 \\
                HiPPO-RNN \citep{gu2020hippo} & 61.10 \\
                LipschitzRNN \citep{erichson2020lipschitz} & 64.20 \\
                \midrule
                \emph{State Space Models}: & \\
                S4 \citep{gu2022parameterization} & 91.80 \\
                S4D \citep{gu2022parameterization} & 90.69 \\
                S5 \citep{smith2023simplified} &  90.10 \\
                Liquid-S4 \citep{hasani2022liquid} & 92.02 \\
                \midrule
                \emph{Convolution}: & \\
                TrellisNet \citep{bai2018trellis} & 73.42 \\
                CKConv \citep{li2022makes} & 63.74 \\
                FlexConv \citep{romero2021flexconv} & 80.82 \\
                MultiresNet \citep{shi2023multiresnet} & \underline{93.15} \\
                {\ours} (ours) & \textbf{94.02} \\
                \bottomrule
            \end{tabular}
        \end{small} 
\end{table}
\begin{table}[b]
    \centering
    \caption{Performance and training speed on WikiText-103 dataset.}
    \label{tab:wiki103}
    \begin{small}
        \begin{tabular}{l|ccc}
        \toprule
                       & \multicolumn{3}{c}{\textbf{WikiText-103}} \\
        \textbf{Model}          & \textbf{\#Param.}         & \textbf{PPL}         & \textbf{Speed}  \\ \hline
        Transformer-adaptive  &247M       &18.66        &5.6 k t/s \\
        Transformer-XL & 257M             & 18.30       & - \\
        S4\citep{gu2020improving}    & 249M             & 20.95      & -   \\
        Mega-chunk\citep{ma2022mega} & 252M             & \underline{18.07}     & 48k t/s    \\
        {\ours} (ours)       & 258M             & \textbf{16.97}       & 53k t/s  \\ \bottomrule
        \end{tabular}
    \end{small}
\end{table}

\begin{table}[ht]
    \centering
    \caption{Testing bits-per-byte on Enwik8 dataset.}
    \label{tab:enwiki}
    \begin{small}
         \begin{tabular}{l|cc}
        \toprule
                       & \multicolumn{2}{c}{\textbf{enwik8}} \\
        \textbf{Model}          & \textbf{\#Param.}       & \textbf{PPL}       \\ \hline
        Transformer-XL & 41M           & 1.06      \\
        Mega~\citep{ma2022mega}           & 39M           & 1.02      \\
        Transformer-VQ~\citep{lingle2023transformervq} & 190M           & \underline{0.99}      \\
        {\ours} (ours)         & 48M           & \textbf{0.96}      \\ \bottomrule
        \end{tabular}
    \end{small}
\end{table}

\subsection{Auto-Regressive Language Modeling}
By following~\citet{ma2022mega,lingle2023transformervq}, we assess {\ours} on two popular language modeling datasets, \textit{i.e.}, WikiText-103~\citep{merity2016wiki103} and enwik8~\citep{2011enwik8}, which are next-token prediction tasks. 
WikiText-103 is a dataset for word-level language modeling with 103 million tokens from Wikipedia articles in its training set. In line with previous work~\citep{baevski2018adaptive}, our method involves using adaptive softmax and input embeddings, and we utilize a vocabulary of 260,000 tokens. Enwik8 is a commonly used benchmark for character-level language modeling, presenting a significant challenge to models. It comprises approximately 100 million unprocessed tokens from Wikipedia articles and has a vocabulary size of about 200. When evaluating language models, we segment the test data and process each segment sequentially during testing to assess their effectiveness.
In Table~\ref{tab:wiki103}, we compare with previous top-performing models that are designed to take advantage of longer context, including Transformers~\citep{baevski2018adaptive}, Transformer-XL and S4~\citep{gu2021efficiently}. 
The model we developed demonstrated outstanding performance on both WikiText-103 and enwik8 datasets, outperforming the baseline models by a significant margin. Our model achieves an inference speed that is almost 10 times faster than the Pure Transformer model. The hybrid structure of the short-long convolutions layer plays a crucial role in enabling our model to manage length extrapolation during inference, allowing it to process longer sequences than those encountered during training. This distinctive characteristic of our model enhances its capability to naturally handle complex tasks, making it a valuable addition to any long-sequence project.

\subsection{Pixel-Level Sequential Image Classification}
Begin by addressing tasks related to image classification, in which images are considered as a one-dimensional sequence of pixels. In these tasks, models cannot rely on preconceived two-dimensional structures within the image. Consequently, the model must possess the ability to recognize patterns at different temporal scales, including pixels that are close together in the original image but far apart in their sequential representation.
We evaluate the performance of our model using the Sequential CIFAR-10 dataset, commonly used as a benchmark for capturing long-term dependencies in RNNs. The CIFAR-10 dataset is frequently employed in machine learning for tasks on image classification. Within this dataset, the typical training and testing split is maintained, reserving 10\% of the training set for validation purposes. To categorize the images, the mean of all tokens in the output sequences is computed, and the resulting values are subjected to a fully connected layer to produce class logits.
The Table~\ref{tab:scifar} displays the results. {\ours} has achieved state-of-the-art performance and the best test accuracy on the sequence classification task, surpassing multiple strong competitors such as Transformers~\citep{vaswani2017attention}, RNNs, state space models, and other convolutional models. In particular, the {\ours} model has exceeded the performance of previous convolution-based models by more than ten percentage points.
It is important to note that our model has delivered impressive results by surpassing the previously established performance standard, even though it uses a relatively simple architecture. The model primarily employs a hybrid method that compresses long historical information based on the output of short-long convolutions. Our most effective model consists of ten {\ours} blocks, which significantly contribute to achieving exceptional performance.

\begin{figure}[b!]
    \centering
    \includegraphics[width=1.0\linewidth]{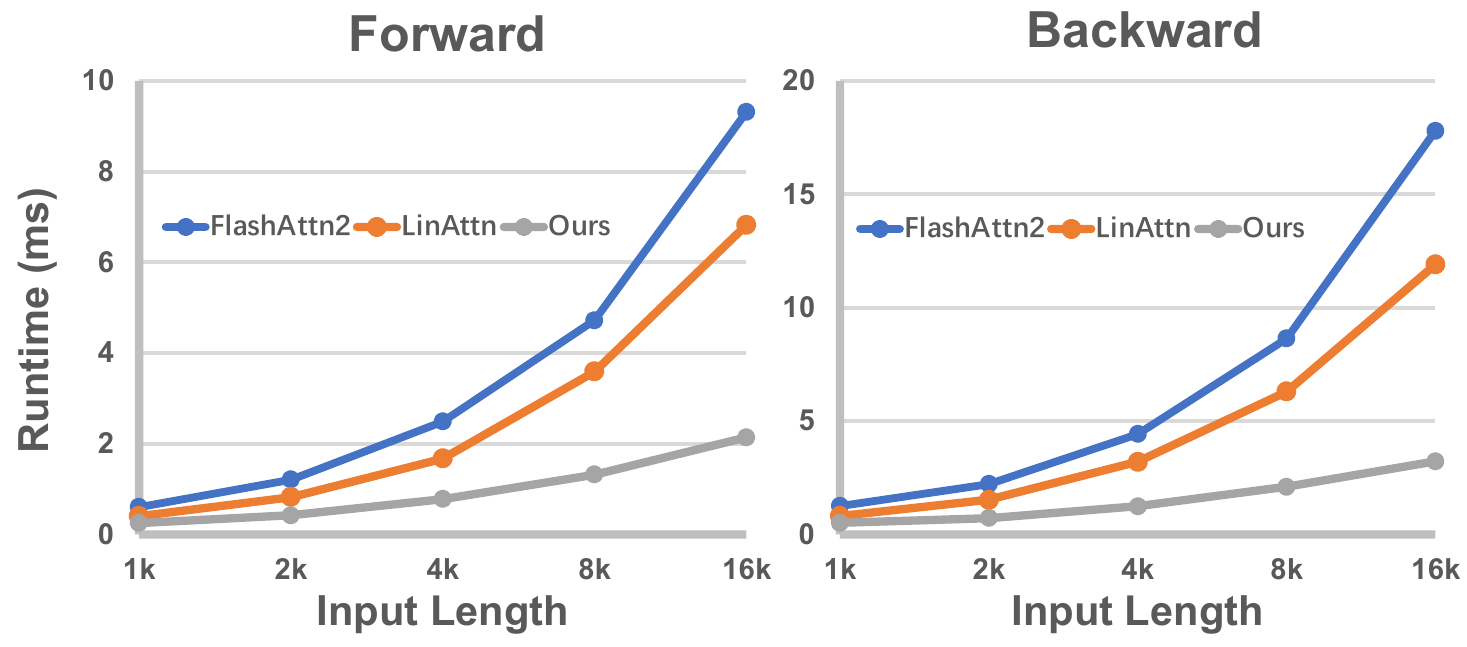}
    \caption{Comparative Analysis of Speed: Runtime in milliseconds for the forward and backward pass across varying lengths.}
    \label{fig:benchmark}
\end{figure}

\section{Ablation Study}
Our ablation experiments focus on answering two key questions mostly related to our design:
(1) Does the hardware-friendly implementation significantly improve the speed of linear attention? 
(2) The effectiveness of our proposed short-long convolutions module on long sequences.

\paragraph{Q1: Benchmark hardware-efficient linear attention.}
To answer the first question, our Hardware-Efficient Linear Attention achieves almost real linear relationships with sequence lengths. We conducted an analysis on the WikiText-103 dataset with models with 200M parameters. As visualized in Fig.~\ref{fig:benchmark}, we have more than doubled the speedup of the original Pytorch implementation of the linear attention.

\paragraph{Q2: Analysis of short-long convolutions.}
To answer the second question, we further combined a variety of hybrid models following the modeling structure of {\ours}. Specifically, we compared the representative SSM-like modules on the subset of the LRA~\citep{tay2020long} dataset (Text, Image, and PathX). It is clear that the proposed Short-Long Convolutions are the best partner for linear attention.

\begin{table}[h]
    \caption{Ablation study on different structured mixers in {\ours}.}
    \centering
    \begin{small}
        \begin{tabular}{lccc}
        \toprule
        \multirow{2}{*}{\textbf{Methods}} & \multicolumn{3}{c}{\textbf{Datasets}}                                                     \\ \cline{2-4} 
                                 & \multicolumn{1}{c}{\textbf{Text}} & \multicolumn{1}{c}{\textbf{Image}} & \multicolumn{1}{c}{\textbf{PathX}} \\ \hline
        Damped EMA~\citep{ma2022mega}               & 90.19                    & 85.80                     & 93.81 \\
        S4D~\citep{gu2022parameterization}                      & 90.85                    & 88.95                     & 94.29 \\
        Long Conv~\cite{fu2023longconv}                & 90.35                    & 87.57                     & 97.24 \\
        Short-Long Convs          & 91.10                    & 91.12                     & 98.65 \\ \bottomrule
        \end{tabular}
    \end{small}
\end{table}
\section{Related Works}

\paragraph{Efficient transformer models}
A variety of efforts have been made to decrease the quadratic time and space complexity of standard attention mechanisms. One method is to utilize ``sparse attention," where each token only attends to a subset of all the tokens based on predefined patterns, such as neighboring tokens within a fixed-size window. \cite{child2019generating} started the attempt to sparse the attention, and then there were a lot more followers, such as 
ETC~\citep{ainslie2020etc}, 
Longformer~\citep{beltagy2020longformer}, BigBird~\citep{zaheer2020big}, Poolingformer~\citep{zhang2021poolingformer}, and HEPOS~\citep{huang2021efficient} are some examples of this approach. Another option is to utilize ``low-rank projection," as mentioned in the work by \cite{wang2020linformer}. Similar techniques include Nyströmformer~\citep{xiong2021nystromformer}, Synthesizer~\citep{tay2021synthesizer}, and Luna ~\citep{ma2021luna}. However, these methods encounter challenges when dealing with causal tasks, such as auto-regressive language modeling.
Another approach uses ``clustering method," where we partition $\Qb$ or $\Kb$ into multiple clusters and perform inter-cluster attention. Examples of such methods include Sinkhorn Transformer~\citep{tay2020sparse}, Reformer~\citep{kitaev2020reformer},  Routing Transformer~\citep{roy2021efficient}, and simplified FLASH~\citep{hua2022transformer}, etc.
``Methods based on kernels" can be utilized to approximate the complete attention $\mathrm{Attn}(\Xb)$. These methods replace the quadratic-time softmax attention with fast linear-time kernel approximations (such as Gaussian and arc-cosine kernels). Some instances of this approach include Linear Transformer~\citep{katharopoulos2020transformers}, Performer~\citep{choromanski2020rethinking}, and FMMformer~\citep{nguyen2021fmmformer}, etc.
Both low-dimensional projection and methods based on kernels are employed to estimate full attention and, as a result, are vulnerable to significant approximation errors.

\paragraph{State space models and long convolutions}
Recurrent neural networks and their linear counterparts such as state-space models are capable of retaining memory of the past. Among these models, S4~\citep{gu2021efficiently} is notable because it can be implemented through convolutions thanks to its linear recurrence. However, the long convolution kernel for this model is as long as the input sequence, and its efficient computation requires sophisticated parameterization and approximation techniques. Although recent advances have found solutions to this issue, initializing these models still requires special effort~\citep{gupta2022diagonal,gu2020improving}. Many of these models use the HiPPO~\cite{gu2020hippo} initialization mechanism, which aims to memorize historical data through projection to orthogonal polynomials. Based on a structure similar to SSM, an increasing number of models focusing on either linear recurrence or global convolution have been developed recently~\citep{fu2023longconv,fu2023h3,poli2023hyena,gu2023mamba}.


\paragraph{Hardware-efficient implementation}
The FlashAttention series~\citep{dao2022flashattention1, dao2023flashattention2} is dedicated to optimizing the standard attention operator for GPU platforms at the system level. Its effectiveness has been extensively validated. The approach involves using tiling strategies to minimize the amount of memory reads/writes between the high bandwidth memory (HBM) and on-chip SRAM.
\section{Conclusion and Limitations}

We presented {\ours} for robust and efficient modeling of long sequences. {\ours} is an SSM-attention hybrid architecture that computes both structured and data-dependent patterns in linear time with respect to sequence length. Its superior performance is enabled by considering the global view of the Short-Long Convolutions and real linear Hardware-Efficient Linear Attention with gating mechanisms. Compared to other hybrid linear models, we are currently the first to achieve performance improvement while maintaining linear complexity. Our large-scale and diverse experiments demonstrate that {\ours} is an efficient and flexible long sequence model with excellent performance on image, text, logical reasoning, and speech data.

However, {\ours} also has some limitations. The optimal combinations of short convolutions are not explored in this paper, which should be a future research direction of designing dynamic short-convolutional components according to the input data. Moreover, the time-varying SSM is a different idea to achieve this goal. By embracing {\ours} as a starting point in the integration of hardware-efficient implementation into hybrid models, we are taking the initial step towards achieving greater efficiency gains in the future.

\section*{Acknowledgements}
This work was supported by Ministry of Science and Technology of the People's Republic of China (No. 2021YFA1301603), National Natural Science Foundation of China Project (No. U21A20427), Project (No. WU2022A009) from the Center of Synthetic Biology and Integrated Bioengineering of Westlake University and Project (No. WU2023C019) from the Westlake University Industries of the Future Research Funding. This work was done when Li Wang and Zedong Wang interned at Westlake University. We thank the AI Station of Westlake University for the support of GPUs.

\section*{Impact Statement}
The goal of this paper is to advance research in long-sequence modeling by introducing an efficient model design {\ours}. We have considered broader ethical impacts and do not foresee {\ours} directly leading to negative societal consequences. All datasets and models used are existing public resources that do not contain private or sensitive information. 
Through discussing the hybrid design of linear attention and SSM models, we aim to make sequence mixers much more efficient by fully leveraging hardware and different model features. Besides, as the community proposes new methods, we encourage discussing any potential negative impacts early in the research process. Overall, we believe hardware-efficient style and the combination of structured and data-dependent patterns are the dominant trends of the future for efficient sequence modeling.

\newpage
\nocite{langley00}

\bibliography{example_paper}
\bibliographystyle{icml2024}

\newpage
\appendix
\onecolumn
\section{Experimental Details}
\label{appendx:exp}
\subsection{Long Range Arena (LRA) and sCIFAR}
For all tasks, we closely follow \citet{tay2020long} for details such as data preprocessing, data split, etc. The hyper-parameters of {\ours} models on these tasks are listed in Table~\ref{tab:lra:hyps}. The experimental configuration of sCIFAR follows the parameter settings of the image in LRA.

\begin{table}[!h]
\caption{Hyper-parameters of {\ours} models on LRA and raw speech classification tasks. BSZ is batch size, LR is learning rate and WD is weight decay.
BN, LN and SN refer to Batch Normalization, Layer Normalization and Scale Normalization.}
\label{tab:lra:hyps}
\centering
\resizebox{\columnwidth}{!}{
    \begin{tabular}{l|ccccccccccc}
    \toprule
    \textbf{Task}       & \textbf{Depth} & $d_\mathrm{model}$ & $d_\mathrm{FFN}$ & \textbf{Attn-FN} & \textbf{Norm} & \textbf{Pre-norm} & \textbf{BSZ} & \textbf{LR} & \textbf{Dropout} & \textbf{WD} & \textbf{Epochs} \\ \hline
    \textbf{ListOps}    & 6                               & 80                 & 160              & laplace                           & BN                             & False                              & 64                            & 0.001                        & 0.1                               & 0.01                         & 60                               \\
    \textbf{Text}       & 4                               & 128                & 256              & norm                           & SN                             & False                              & 50                            & 0.004                        & 0.1                               & 0.01                         & 50                               \\
    \textbf{Retrieval}  & 6                               & 128                & 256              & norm                           & SN                             & False                              & 64                            & 0.003                        & 0.1                               & 0.04                         & 40                               \\
    \textbf{Image}      & 8                               & 160                & 320              & laplace                           & BN                             & True                               & 50                            & 0.01                         & 0.0                               & 0.02                         & 200                              \\
    \textbf{Pathfinder} & 6                               & 128                & 256              & laplace                           & BN                             & True                               & 128                           & 0.01                         & 0.0                               & 0.01                         & 200                              \\
    \textbf{Path-X}     & 4                               & 64                 & 128              & laplace                           & BN                             & True                               & 128                           & 0.01                         & 0.0                               & 0.01                         & 100                              \\ \hline
    \textbf{SC}         & 6                               & 60                 & 120              & laplace                           & BN                             & True                               & 20                            & 0.01                         & 0.0                               & 0.01                         & 200   \\     \bottomrule                        
    \end{tabular}
}
\end{table}

\subsection{Language Modeling}
We use the data of WikiText-103 and enwik8 and their splits provided by \citet{ma2022mega}. At training time, we split the training data into segments; each segment contains $m$ consecutive chunks, where the chunk size is the effective attention length. Other training hyperparameters, including optimizer, learning rate scheduler, and architecture, are presented in Table~\ref{tab:lm:hyps}.

\begin{table}[!h]
\caption{Hyper-parameters of models for language modeling.}
\label{tab:lm:hyps}
\centering
    \begin{tabular}{l|ccc}
    \toprule
     & \textbf{WikiText-103}  & \textbf{enwik8} \\
    \midrule
    Batch Size $\times$ GPUs  & 6144 $\times$ 24 & 8192 $\times$ 8 \\
    Optimizer & AdamW & AdamW \\
    Learning Rate &  0.005 & 0.005 \\
    Adam-$\beta$ & $(0.9, 0.98)$ & $(0.9, 0.98)$ \\
    Learning Rate Decay & linear & linear \\
    Weight Decay & 0.1 & 0.1 \\
    Dropout & 0.3 & 0.1 \\
    Attention Dropout & 0.1 & 0.0 \\
    FFN Hidden Dropout & 0.1 & 0.0 \\
    Gradient Clipping & 1.0 & 1.0 \\
    Warmup steps & 24K & 24K \\
    Total updates & 400K & 400K \\
    \midrule
    Decoder Layers & 16 & 12 \\
    Model size & 1024 & 512 \\
    FFN Hidden size & 1536 & 1024 \\
    Total Parameters & 258M & 48M \\
    \bottomrule
    \end{tabular}
\end{table}


\end{document}